\documentclass{article}

\usepackage[accepted]{icml2021}

\usepackage{pbox}
\usepackage[utf8]{inputenc} 
\usepackage{hyperref}       
\usepackage{url}            
\usepackage{booktabs}       
\usepackage{multirow}
\usepackage{amsfonts}       
\usepackage{amsmath}
\usepackage{amsthm}
\usepackage{amssymb}
\usepackage{graphicx,wrapfig,lipsum}
\usepackage{empheq}
\usepackage{mathtools}
\usepackage{mathrsfs}  
\usepackage{tikz}
\usepackage{nicefrac}
\usepackage{bbm}
\usepackage{arydshln}
\usepackage{xfrac}
\usepackage[shortlabels]{enumitem}
\usetikzlibrary{fit, calc,positioning,matrix, decorations.pathreplacing}
\usepackage{verbatim}
\usepackage{dblfloatfix}

\newcommand{\expect}{\mathbb{E}}

\newcommand{\abs}[1]{\left|#1\right|}
\newcommand{\reals}{\mathbb{R}}

\newcommand{\bigO}{\mathcal{O}}

\newcommand{\dd}{\mathrm{d}}
\newcommand{\normal}[2]{\mathcal{N}\left(#1, #2\right)}

\newcommand{\restr}[2]{{\left.\kern-\nulldelimiterspace #1 \right|_{#2}}}

\definecolor{mydarkblue}{rgb}{0,0.08,0.45}
\hypersetup{
    colorlinks=true,
    linkcolor=mydarkblue,
    citecolor=mydarkblue,
    filecolor=mydarkblue,
    urlcolor=mydarkblue
}

\begin{document}

\twocolumn[
\icmltitle{Neural SDEs as Infinite-Dimensional GANs}

\begin{icmlauthorlist}
\icmlauthor{Patrick Kidger}{oxford,ati}
\icmlauthor{James Foster}{oxford,ati}
\icmlauthor{Xuechen Li}{stanford}
\icmlauthor{Harald Oberhauser}{oxford,ati}
\icmlauthor{Terry Lyons}{oxford,ati}
\end{icmlauthorlist}

\icmlaffiliation{oxford}{Mathematical Institute, University of Oxford}
\icmlaffiliation{ati}{The Alan Turing Institute, The British Library}
\icmlaffiliation{stanford}{Stanford}

\icmlcorrespondingauthor{Patrick Kidger}{kidger@maths.ox.ac.uk}

\icmlkeywords{machine learning, deep learning, neural differential equations, neural sdes, neural odes, gan, generative}

\vskip 0.3in
]
\printAffiliationsAndNotice{}

\begin{abstract}
Stochastic differential equations (SDEs) are a staple of mathematical modelling of temporal dynamics. However, a fundamental limitation has been that such models have typically been relatively inflexible, which recent work introducing Neural SDEs has sought to solve. Here, we show that the current classical approach to fitting SDEs may be approached as a special case of (Wasserstein) GANs, and in doing so the neural and classical regimes may be brought together. The input noise is Brownian motion, the output samples are time-evolving paths produced by a numerical solver, and by parameterising a discriminator as a Neural Controlled Differential Equation (CDE), we obtain Neural SDEs as (in modern machine learning parlance) continuous-time generative time series models. Unlike previous work on this problem, this is a direct extension of the classical approach without reference to either prespecified statistics or density functions. Arbitrary drift and diffusions are admissible, so as the Wasserstein loss has a unique global minima, in the infinite data limit any SDE may be learnt. Example code has been made available as part of the \texttt{torchsde} repository.
\end{abstract}

\section{Introduction}
\subsection{Neural differential equations}
Since their introduction, neural ordinary differential equations \citep{neural-odes} have prompted the creation of a variety of similarly-inspired models, for example based around controlled differential equations \citep{kidger2020neuralcde, morrill2020logode}, Lagrangians \citep{cranmer2020lagrangian}, higher-order ODEs \citep{dissecting, sonode}, and equilibrium points \citep{deq}.

In particular, several authors have introduced \emph{neural stochastic differential equations} (neural SDEs), such as \citet{nsde-basic, scalable-sde, roughstochasticNF}, among others. This is our focus here.

Neural differential equations parameterise the vector field(s) of a differential equation by neural networks. They are an elegant concept, bringing together the two dominant modelling paradigms of neural networks and differential equations.

The main idea -- fitting a parameterised differential equation to data, often via stochastic gradient descent -- has been a cornerstone of mathematical modelling for a long time \citep{smoking-adjoints}. The key benefit of the neural network hybridisation is its availability of easily-trainable high-capacity function approximators.

\subsection{Stochastic differential equations}
Stochastic differential equations have seen widespread use for modelling real-world random phenomena, such as particle systems \citep{langevinbook, langevinbook2, sdesformoldynamics}, financial markets \citep{blackscholes, cir, ratebook}, population dynamics \citep{stocLotkaVolterra, populationgrowth} and genetics \citep{wrightfisher}. They are a natural extension of ordinary differential equations (ODEs) for modelling systems that evolve in continuous time subject to uncertainty.

The dynamics of an SDE consist of a deterministic term and a stochastic term:
\begin{equation}\label{eq:sde}
\dd X_t = f(t, X_t) \,\dd t + g(t, X_t) \circ \dd W_t,
\end{equation}
where $X = \{X_t\}_{t\in[0,T]}$ is a continuous $\reals^x$-valued stochastic process, $f:[0,T]\times \reals^x\to \reals^x$, $g:[0,T]\times \reals^x\to\reals^{x\times w}$ are functions and $W = \{W_t\}_{t\geq 0}$ is a $w$-dimensional Brownian motion. We refer the reader to \citet{bbridge} for a rigorous account of stochastic integration.

\newcommand{\Brownian}[9]{
\pgfmathsetseed{#9}
\draw[#4] (#5, #6)
\foreach \x in {1,...,#1}
{   -- ++(#2,#8 + rand*#3)
}
node[right] {#7};
}

\newcommand{\SDE}[9]{
\pgfmathsetseed{#9}
\draw[#4] (#5, #6)
\foreach \x in {1,...,#1}
{   -- ++(#2,#8*\x - #8*0.5*#1 + rand*#3)
}
node[right] {#7};
}

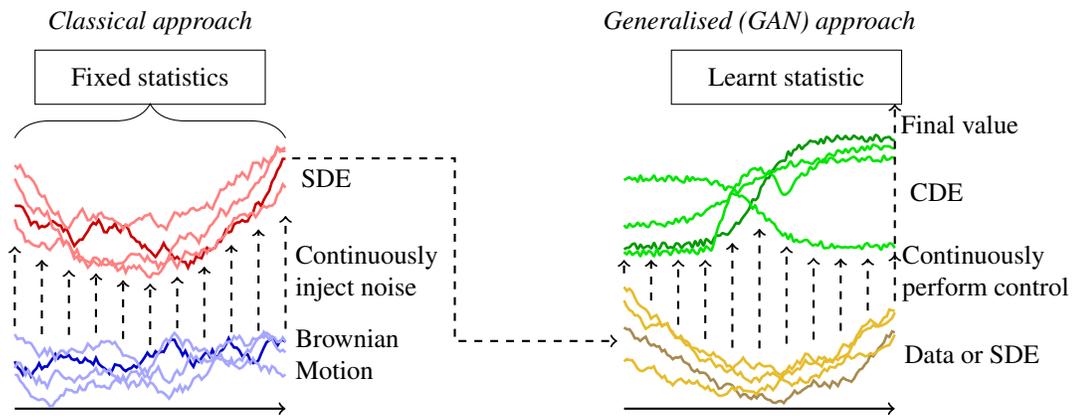
\begin{figure*}[!b]\centering
\begin{tikzpicture}[scale=0.9]
    \node at (-1.5,0) {};
    
    \draw[->, thick] (0, 0) -- (4, 0);
    \draw[->, thick] (9, 0) -- (13, 0);
    
    \Brownian{100}{0.04}{0.1}{color=blue!80!black, text=black, line width=1pt}{0}{0.7}{}{0.005}{56791}
    \Brownian{100}{0.04}{0.1}{color=blue!35!white, line width=1pt}{0}{0.6}{}{0.005}{56789}
    \Brownian{100}{0.04}{0.1}{color=blue!35!white, line width=1pt}{0}{0.5}{}{0.005}{56790}
    \Brownian{100}{0.04}{0.1}{color=blue!35!white, line width=1pt}{0}{1.1}{}{0.005}{45678}
    
    \node at (5.1, 0.8) {\begin{minipage}{0.1\textwidth}Brownian\\Motion\end{minipage}};
    
    \SDE{100}{0.04}{0.1}{color=red!80!black, text=black, line width=1pt}{0}{3}{}{0.0008}{7890}
    \SDE{100}{0.04}{0.1}{color=red!50!white, line width=1pt}{0}{3.4}{}{0.0008}{7891}
    \SDE{100}{0.04}{0.1}{color=red!50!white, line width=1pt}{0}{3.6}{}{0.0008}{7892}
    \SDE{100}{0.04}{0.1}{color=red!50!white, line width=1pt}{0}{2.8}{}{0.0008}{7893}
    
    \node at (4.6, 3.4) {SDE};
    
    \draw[decorate, decoration={brace, amplitude=13pt}] (0, 4) -- (4, 4);
    
    \draw (0.3,4.5) rectangle (3.7,5.3) node[midway] {Fixed statistics};
    
    \node at (2, 5.7) {\textit{Classical approach}};
    
    \begin{scope}[shift={(1,0)}]
    \SDE{100}{0.04}{0.07}{color=yellow!70!red!60!black, text=black, line width=1pt}{8}{1.2}{}{0.0008}{7894}
    \SDE{100}{0.04}{0.07}{color=yellow!80!red!90!black, line width=1pt}{8}{1.6}{}{0.0008}{7895}
    \SDE{100}{0.04}{0.07}{color=yellow!80!red!90!black, line width=1pt}{8}{1.8}{}{0.0008}{7896}
    \SDE{100}{0.04}{0.07}{color=yellow!80!red!90!black, line width=1pt}{8}{0.7}{}{0.0006}{7897}
    
    \node at (13.14, 0.8) {Data or SDE};
    
    \begin{scope}[shift={(10, 3)}]
    \draw[color=green!60!black, line width=1pt] plot [domain=-2:2, samples=100, smooth] (\x, {0.2 + 0.8 * tanh(\x*2) + 0.05*rand});
    
    \draw[color=green!90!black, line width=1pt] plot [domain=-2:2, samples=100, smooth] (\x, {0.2 + 0.5 * tanh(\x*1.5 + 0.5) + 0.05*rand});
    
    \draw[color=green!90!black, line width=1pt] plot [domain=-2:2, samples=100, smooth] (\x, {0.1 + max(0,0.8-3*\x*\x-\x) + 0.8 * tanh(\x*1.5 - 0.5) + 0.05*rand});
    
    \draw[color=green!90!black, line width=1pt] plot [domain=-2:2, samples=100, smooth] (\x, {-0.1 - 0.5 * tanh(\x*2) + 0.05*rand});
    
    \node at (2.62, 0.2) {CDE};
    
    \draw[dashed, line width=0.8pt, ->] (2, -0.8) -- (2, 1.5);
    
    \node at (2.96, 1.2) {Final value};
    
    \draw (-1.3,1.5) rectangle (2.1,2.3) node[midway] {Learnt statistic};
    
    \node at (0, 2.7) {\textit{Generalised (GAN) approach}};
    
    \end{scope}
    \draw[dashed, line width=0.8pt, ->] (8,2) -- (8,2.2);
    \draw[dashed, line width=0.8pt, ->] (8.4,1.6) -- (8.4,2.2);
    \draw[dashed, line width=0.8pt, ->] (8.8,1.24) -- (8.8,2.2);
    \draw[dashed, line width=0.8pt, ->] (9.2,1.03) -- (9.2,2.2);
    \draw[dashed, line width=0.8pt, ->] (9.6,0.9) -- (9.6,2.45);
    \draw[dashed, line width=0.8pt, ->] (10,0.89) -- (10,2.66);
    \draw[dashed, line width=0.8pt, ->] (10.4,0.96) -- (10.4,2.4);
    \draw[dashed, line width=0.8pt, ->] (10.8,1.02) -- (10.8,2.23);
    \draw[dashed, line width=0.8pt, ->] (11.2,1.15) -- (11.2,2.22);
    \draw[dashed, line width=0.8pt, ->] (11.6,1.45) -- (11.6,2.22);
    \draw[dashed, line width=0.8pt, ->] (12,1.59) -- (12,2.28);
    \node at (13.35, 2) {\begin{minipage}{0.13\textwidth} Continuously perform control\end{minipage}};
    \end{scope}
    
    \draw[dashed, line width=0.8pt, ->] (4.1, 3.7) -- (6.5, 3.7) -- (6.5, 1) -- (8.9, 1);
    
    \draw[dashed, line width=0.8pt, ->] (0,1.22) -- (0,2.4);
    \draw[dashed, line width=0.8pt, ->] (0.4,1.1) -- (0.4,2.2);
    \draw[dashed, line width=0.8pt, ->] (0.8,1.09) -- (0.8,2.05);
    \draw[dashed, line width=0.8pt, ->] (1.2,1.12) -- (1.2,2);
    \draw[dashed, line width=0.8pt, ->] (1.6,1.03) -- (1.6,1.9);
    \draw[dashed, line width=0.8pt, ->] (2,0.94) -- (2,1.85);
    \draw[dashed, line width=0.8pt, ->] (2.4,1.19) -- (2.4,1.95);
    \draw[dashed, line width=0.8pt, ->] (2.8,1.05) -- (2.8,2.1);
    \draw[dashed, line width=0.8pt, ->] (3.2,1.13) -- (3.2,2.42);
    \draw[dashed, line width=0.8pt, ->] (3.6,1.15) -- (3.6,2.65);
    \draw[dashed, line width=0.8pt, ->] (4,1.18) -- (4,2.85);
    \node at (5.1, 2) {\begin{minipage}{0.1\textwidth} Continuously inject noise\end{minipage}};
\end{tikzpicture}
\vspace{-0.5em}
\caption{Pictorial summary of just the high level ideas: Brownian motion is continuously injected as noise into an SDE. The classical approach fits the SDE to prespecified statistics. Generalising to (Wasserstein) GANs, which instead introduce a learnt statistic (the discriminator), we may fit much more complicated models.}
\end{figure*}

The notation ``$\circ$'' in the noise refers to the SDE being understood using Stratonovich integration. The difference between It{\^o} and Stratonovich will not be an important choice here; we happen to prefer the Stratonovich formulation as the dynamics of (\ref{eq:sde}) may then be informally interpreted as
\begin{equation*}
X_{t+\Delta t} \approx \text{ODESolve}\bigg(X_t\,, f(\,\cdot\,) + g(\,\cdot\,)\frac{\Delta W}{\Delta t}\,,\, [t, t + \Delta t]\bigg),
\end{equation*}
where $\Delta W\sim\normal{0}{\Delta t I_w}$ denotes the increment of the Brownian motion over the small time interval $[t, t+\Delta t]$.

Historically, workflows for SDE modelling have two steps:
\begin{enumerate}[topsep=0pt]
\item A domain expert will formulate an SDE model using their experience and knowledge. One frequent and straightforward technique is to add ``$\sigma \circ \dd W_t$'' to a pre-existing ODE model, where $\sigma$ is a fixed matrix.

\item Once an SDE model is chosen, the model parameters must be calibrated from real-world data. Since SDEs produce random sample paths, parameters are often chosen to capture some desired expected behaviours. That is, one trains the model to match target statistics:
\begin{equation}\label{eq:sde_stats}
\big\{\expect\big[F_i(X)\big]\big\}_{1\leq i\leq n},
\end{equation}
where the real-valued functions $\{F_i\}$ are prespecified.
For example in mathematical finance, the statistics (\ref{eq:sde_stats}) represent option prices that correspond to the functions $F_i$, which are termed payoff functions; for the well-known and analytically tractable Black--Scholes model, these prices can then be computed explicitly for call and put options \citep{blackscholes}.


\end{enumerate}
The aim of this paper (and neural SDEs more generally) is to strengthen the capabilities of SDE modelling by hybridising with deep learning. 

\subsection{Contributions}
SDEs are a classical way to understand uncertainty over paths or over time series. Here, we show that the current classical approach to fitting SDEs may be generalised, and approached from the perspective of Wasserstein GANs. In particular this is done by putting together a neural SDE and a neural CDE (controlled differential equation) as a generator--discriminator pair.

Arbitrary drift and diffusions are admissible, which from the point of view of the classical SDE literature offers unprecedented modelling capacity. As the Wasserstein loss has a unique global minima, then in the infinite data limit arbitrary SDEs may be learnt.

Unlike much previous work on neural SDEs, this operates as a direct extension of the classical tried-and-tested approach. Moreover and to the best of our knowledge, this is the first approach to SDE modelling that involves neither prespecified statistics nor the use of density functions.

In modern machine learning parlance, neural SDEs become continuous-time generative models. We anticipate applications in the main settings for which SDEs are already used -- now with enhanced modelling power. For example later we will consider an application to financial time series.

\begin{figure*}[t]
    \centering
    \newcommand{\fillcolour}{yellow!40!white}
\newcommand{\backfillcolour}{green!10!white}

\begin{tikzpicture}
    \draw[gray, dashed, fill=\backfillcolour] (-2.5, -0.1) rectangle ++(14.1, 0.8);
    \draw[gray, dashed, fill=\backfillcolour] (-2.5, 1.1) rectangle ++(14.1, 0.8);
    \draw[gray, dashed, fill=\backfillcolour] (-2.5, 2.1) rectangle ++(11.05, 0.8);

    \draw[rounded corners, fill=\fillcolour] (0, 0) rectangle node {$H_0 = \xi_\phi(Y_0)$} ++(2.3, 0.6);
    
    \draw[rounded corners, fill=\fillcolour] (0, 1.2) rectangle node {$X_0 = \zeta_\theta(V)$} ++(2.3, 0.6);
    \draw[rounded corners, fill=\fillcolour] (0, 2.2) rectangle node {$V \sim \normal{0}{I_v}$} ++(2.3, 0.6);
    
    \draw[rounded corners, fill=\fillcolour] (2.75, 2.2) rectangle node {$W_t =$ Brownian motion} ++(5.7, 0.6);
    \draw[rounded corners, fill=\fillcolour] (2.75, 1.2) rectangle node {$\dd X_t = \mu_\theta(t, X_t) \,\dd t + \sigma_\theta(t, X_t) \circ \dd W_t$} ++(5.7, 0.6);
    \draw[rounded corners, fill=\fillcolour] (2.75, 0) rectangle node {$\dd H_t = f_\phi(t, H_t) \,\dd t + g_\phi(t, H_t) \circ \dd Y_t$} ++(5.7, 0.6);
    
    \draw[rounded corners, fill=\fillcolour] (8.9, 0) rectangle node {$D = m_\phi \cdot H_T$} ++(2.6, 0.6);
    
    \draw[rounded corners, fill=\fillcolour] (8.9, 1.2) rectangle node {$Y_t = \alpha_\theta X_t + \beta_\theta$} ++(2.6, 0.6);
    
    \draw[->, thick] (2.3, 1.5) -- (2.75, 1.5);
    \draw[->, thick] (2.3, 0.3) -- (2.75, 0.3);
    
    \draw[->, thick] (8.45, 1.5) -- (8.9, 1.5);
    \draw[->, thick] (8.45, 0.3) -- (8.9, 0.3);
    
    \draw[->, thick] (10.05, 1.2) -- (10.05, 0.9) -- (1.15, 0.9) -- (1.15, 0.6);

    \draw[->, thick] (5.6, 2.2) -- (5.6, 1.8);
    
    \draw[->, thick] (5.6, 0.9) -- (5.6, 0.6);
        
    \draw[->, thick] (1.15, 2.2) -- (1.15, 1.8);
        
    \draw (-1.3, 2.5) node[label=center:\textbf{Noise}] {};
    \draw (-1.3, 1.5) node[label=center:\textbf{Generator}] {};
    \draw (-1.25, 0.3) node[label=center:\textbf{Discriminator}] {};
    
    \draw (1.15, 3.2) node[label=center:\textbf{Initial}] {};
    \draw (5.6, 3.2) node[label=center:\textbf{Hidden state}] {};
    \draw (10.2, 3.2) node[label=center:\textbf{Output}] {};
\end{tikzpicture}
    \caption{Summary of equations.}\label{fig:summary}
\end{figure*}
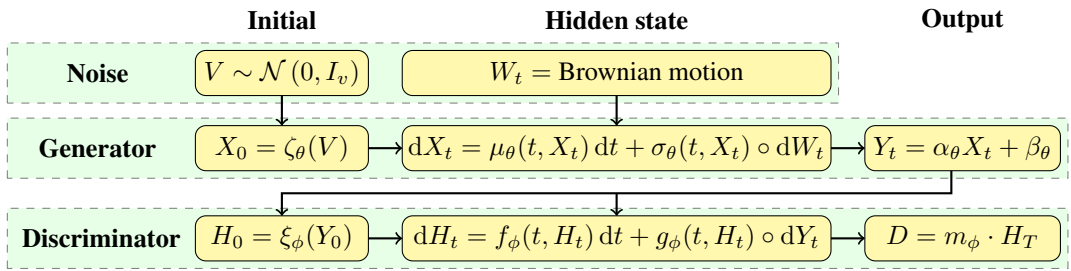

\section{Related work}
We begin by discussing previous formulations, and applications, of neural SDEs. Broadly speaking these may be categorised in two groups. The first use SDEs as a way to gradually insert noise into a system, so that the terminal state of the SDE is the quantity of interest. The second instead consider the full time-evolution of the SDE as the quantity of interest.

\citet{nsde-basic, nsde-generative} obtain Neural SDEs as a continuous limit of deep latent Gaussian models. They train by optimising a variational bound, using forward-mode autodifferentiation. They consider only theoretical applications, for modelling distributions as the terminal value of an SDE.

\citet{scalable-sde} give arguably the closest analogue to the neural ODEs of \citet{neural-odes}. They introduce neural SDEs via a subtle argument involving two-sided filtrations and backward Stratonovich integrals, but in doing so are able to introduce a backward-in-time adjoint equation, using only efficient-to-compute vector-Jacobian products. In applications, they use neural SDEs in a latent variable modelling framework, using the stochasticity to model Bayesian uncertainty.

\citet{roughstochasticNF} introduce neural SDEs as a limit of random ODEs. The limit is made meaningful via rough path theory. In applications, they use the limiting random ODEs, and treat stochasticity as a regulariser within a normalising flow. However, they remark that in this setting the optimal diffusion is zero. This is a recurring problem: \citet{delta-p} also train neural SDEs for which the optimal diffusion is zero.

\citet{rackauckas2020universal} treat neural SDEs in classical Feynman--Kac fashion, and like \citet{roughstochasticNF, nsde-basic, nsde-generative}, optimise a loss on just the terminal value of the SDE.

\citet{nsde-mmd, finance-nsde, finance-nsde2} instead consider the more general case of using a neural SDE to model a time-varying quantity, that is to say not just considering the terminal value of the SDE. Letting $\mu, \nu$ denote the learnt and true distributions on path space, they all train by minimising $\abs{\int f \mathrm{d}\mu - \int f\mathrm{d}\nu}$ for functions of interest $f$ (such as derivative payoffs). This corresponds to training with a non-characteristic MMD~\citep{mmd}.

Several authors, such as \citet{stochasticnode, roughstochasticNF, nsde-normalisation}, seek to use stochasticity as a way to enhance or regularise a neural ODE model.

\citet{song2021scorebased}, building on the discrete time counterparts \citet{score-sde-discrete1, score-sde-discrete2}, consider an SDE that is fixed (and prespecified) rather than learnt. However by approximating one of its terms with a neural network trained with score matching, then the SDE becomes a controlled way to inject noise so as to sample from complex high-dimensional distributions such as images.

Our approach is most similar to \citet{scalable-sde}, in that we treat neural SDEs as learnt continuous-time model components of a differentiable computation graph. Like both \citet{rackauckas2020universal} and \citet{finance-nsde} we emphasise the connection of our approach to standard mathematical formalisms. In terms of the two groups mentioned at the start of this section, we fall into the second: we use stochasticity to model distributions on path space. The resulting neural SDE is not an improvement to a similar neural ODE, but a standalone concept in its own right.

\section{Method}
\subsection{SDEs as GANs}
Consider some (Stratonovich) integral equation of the form
\begin{equation*}
    X_0 \sim \mu, \quad \dd X_t = f(t, X_t) \,\dd t + g(t, X_t) \circ \dd W_t,
\end{equation*}
for initial probability distribution $\mu$, (Lipschitz continuous) functions $f$, $g$ and Brownian motion $W$. The strong solution to this SDE may be defined as the unique function $S$ such that $S(\mu, W) = X$ almost surely \citep[Chapter V, Definition 10.9]{roger-williams}.

Intuitively, this means that SDEs are maps from a noise distribution (Wiener measure, the distribution of Brownian motion) to some solution distribution, which is a probability distribution on path space.

We recommend any of \citet{karatzas1991brownian}, \citet{roger-williams}, or \citet{bbridge} as an introduction to the theory of SDEs.

SDEs can be sampled from: this is what a numerical SDE solver does. However, evaluating its probability density is not possible; in fact it is not even defined in the usual sense.\footnote{Technically speaking, a probability density is the Radon--Nikodym derivative of the measure with respect to the Lebesgue measure. However, the Lebesgue measure only exists for finite dimensional spaces. In infinite dimensions, it is possible to define densities with respect to for example Gaussian measures, but this is less obviously meaningful when used with maximum likelihood.} As such, an SDE is typically fit to data by asking that the model statistics
\begin{equation*}
\big\{\expect_{X \sim \text{model}}\big[F_i(X)\big]\big\}_{1\leq i\leq n},
\end{equation*}
match the data statistics
\begin{equation*}
\big\{\expect_{X \sim \text{data}}\big[F_i(X)\big]\big\}_{1\leq i\leq n},
\end{equation*}
for some functions of interest $F_i$. Training may be done via stochastic gradient descent \citep{smoking-adjoints}.

For completeness we now additionally introduce the relevant ideas for GANs. Consider some noise distribution $\mu$ on a space $\mathcal{X}$, and a target probability distribution $\nu$ on a space $\mathcal{Y}$. A generative model for $\nu$ is a learnt function $G_\theta \colon \mathcal{X} \to \mathcal{Y}$ trained so that the (pushforward) distribution $G_\theta(\mu)$ approximates $\nu$. Sampling from a trained model is typically straightforward, by sampling $\omega \sim \mu$ and then evaluating $G_\theta(\omega)$.

Many training methods rely on obtaining a probability density for $G_\theta(\mu)$; for example this is used in normalising flows \citep{norm-flows}. However this is not in general computable, perhaps due to the complicated internal structure of $G_\theta$. Instead, GANs examine the statistics of samples from $G_\theta(\mu)$, and seek to match the statistics of the model to the statistics of the data. Most typically this is a learnt scalar statistic, called the discriminator. An optimally-trained generator is one for which
\begin{equation*}
    \expect_{X \sim \text{model}}\big[F(X)\big] = \expect_{X \sim \text{data}}\big[F(X)\big]
\end{equation*}
for all statistics $F$, so that there is no possible statistic (or `witness function' in the language of integral probability metrics \cite{demystifying-mmd-gan}) that the discriminator may learn to represent, so as to distinguish real from fake.

There are some variations on this theme; GMMNs instead use fixed vector-valued statistics \citep{gmmn}, and MMD-GANs use learnt vector-valued statistics \citep{mmd-gan}.

In both cases -- SDEs and GANs -- the model generates samples by transforming random noise. In neither case are densities available. However sampling is available, so that model fitting may be performed by matching statistics. With this connection in hand, we now seek to combine these two approaches.

\subsection{Generator}
Let $Y_{\text{true}}$ be a random variable on $y$-dimensional path space. Loosely speaking, path space is the space of continuous functions $f \colon [0, T] \to \reals^y$ for some fixed time horizon $T > 0$. For example, this may correspond to the (interpolated) evolution of stock prices over time. $Y_{\text{true}}$ is what we wish to model.

Let $W \colon [0, T] \to \reals^w$ be a $w$-dimensional Brownian motion, and $V \sim \normal{0}{I_v}$ be drawn from a $v$-dimensional standard multivariate normal. The values $w, v$ are hyperparameters describing the size of the noise.

Let
\begin{align*}
    \zeta_\theta \colon \reals^v &\to \reals^x,\\
    \mu_\theta \colon [0, T] \times \reals^x &\to \reals^x,\\
    \sigma_\theta \colon [0, T] \times \reals^x &\to \reals^{x \times w},\\
    \alpha_\theta \in\null &\reals^{y \times x}\\
    \beta_\theta \in\null &\reals^y
\end{align*}
where $\zeta_\theta$, $\mu_\theta$ and $\sigma_\theta$ are (Lipschitz) neural networks. Collectively they are parameterised by $\theta$. The dimension $x$ is a hyperparameter describing the size of the hidden state.

We define neural SDEs of the form
\begin{align}\label{eq:nsde}
X_0 & = \zeta_\theta(V),\nonumber\\[2pt]
\dd X_t & = \mu_\theta(t, X_t) \,\dd t + \sigma_\theta(t, X_t) \circ \dd W_t,\\[2pt]
Y_t & = \alpha_\theta X_t + \beta_\theta,\nonumber
\end{align}
for $t \in [0, T]$, with $X \colon [0, T] \to \reals^x$ the (strong) solution to the SDE, such that in some sense $Y \overset{\mathrm{d}}{\approx} Y_{\text{true}}$. That is to say, the model $Y$ should have approximately the same distribution as the target $Y_{\text{true}}$ (for some notion of approximate). The solution $X$ is guaranteed to exist given mild conditions (such as Lipschitz $\mu_\theta$, $\sigma_\theta$).

\paragraph{Architecture}
Equation \eqref{eq:nsde} has a certain minimum amount of structure. First, the solution $X$ represents hidden state. If it were the output, then future evolution would satisfy a Markov property which need not be true in general. This is the reason for the additional readout operation to $Y$. Practically speaking $Y$ may be concatenated alongside $X$ during an SDE solve.

Second, there must be an additional source of noise for the initial condition, passed through a nonlinear $\zeta_\theta$, as $Y_0 = \alpha_\theta \zeta_\theta(V) + \beta_\theta$ does not depend on the Brownian noise $W$.

$\zeta_\theta, \mu_\theta$, and $\sigma_\theta$ may be taken to be any standard network architecture, such as a simple feedforward network. (The choice does not affect the GAN construction.)

\paragraph{Sampling} Given a trained model, we sample from it by sampling some initial noise $V$ and some Brownian motion $W$, and then solving equation \eqref{eq:nsde} with standard numerical SDE solvers. In our experiments we use the midpoint method, which converges to the Stratonovich solution. (The Euler--Maruyama method converges to the It{\^o} solution).

\paragraph{Comparison to the Fokker--Planck equation} The distribution of an SDE, as learnt by a neural SDE, contains more information than the distribution obtained by learning a corresponding Fokker--Planck equation. The solution to a Fokker--Planck equation gives the (time evolution of the) probability density of a solution \emph{at fixed times}. It does not encode information about the time evolution of individual sample paths. This is exemplified by stationary processes, whose sample paths may be nonconstant but whose distribution does not change over time.

\paragraph{Stratonovich versus It{\^o}} The choice of Stratonovich solutions over It{\^o} solutions is not mandatory. As the vector fields are learnt then in general either choice is equally admissible.

\subsection{Discriminator}
Each sample from the generator is a path $Y \colon [0, T] \to \reals^y$; these are infinite dimensional and the discriminator must accept such paths as inputs. There is a natural choice: parameterise the discriminator as another neural SDE.

Let
\begin{align}
\xi_\phi \colon \reals^y &\to \reals^h,\nonumber\\
f_\phi \colon [0, T] \times \reals^h &\to \reals^{h},\nonumber\\
g_\phi \colon [0, T] \times \reals^h &\to \reals^{h \times y},\nonumber\\
m_\phi \in\null &\reals^h\label{eq:discriminator}
\end{align}
where $\xi_\phi$, $f_\phi$ and $g_\phi$ are (Lipschitz) neural networks. Collectively they are parameterised by $\phi$. The dimension $h$ is a hyperparameter describing the size of the hidden state.

Recalling that $Y$ is the generated sample, we take the discriminator to be an SDE of the form
\begin{align}\label{eq:ncde}
    H_0 & = \xi_\phi(Y_0),\nonumber\\[2pt]
    \dd H_t & = f_\phi(t, H_t) \,\dd t + g_\phi(t, H_t) \circ \dd Y_t,\\[2pt]
    \qquad D & = m_\phi \cdot H_T,\nonumber
\end{align}
for $t \in [0, T]$, with $H \colon [0, T] \to \reals^h$ the (strong) solution to this SDE, which exists given mild conditions (such as Lipschitz $f_\phi$, $g_\phi$). The value $D \in \reals$, which is a function of the terminal hidden state $H_T$, is the discriminator's score for real versus fake.

\paragraph{Neural CDEs} The discriminator follows the formulation of a neural CDE \citep{kidger2020neuralcde} with respect to the control $Y$. Neural CDEs are the continuous-time analogue to RNNs, just as neural ODEs are the continuous-time analogue to residual networks \citep{neural-odes}. This is what motivates equation \eqref{eq:ncde} as a probably sensible choice of discriminator. Moreover, it means that the discriminator enjoys properties such as universal approximation.

\paragraph{Architecture} There is a required minimum amount of structure. There must be a learnt initial condition, and the output should be a function of $H_T$ and not a univariate $H_T$ itself. See \citet{kidger2020neuralcde}, who emphasise these points in the context of CDEs specifically.

\paragraph{Single SDE solve} In practice, both generator and discriminator may be concatenated together into a single SDE solve. The state is the combined $[X, H]$, the initial condition is the combined
\begin{equation*}
    [\zeta_\theta(V),\, \xi_\phi(\alpha_\theta \zeta_\theta(V) + \beta_\theta)],
\end{equation*}
the drift is the combined
\begin{equation*}
    [\mu_\theta(t, X_t),\, f_\phi(t, H_t) + g_\phi(t, H_t) \alpha_\theta \mu_\theta(t, X_t)],
\end{equation*}
and the diffusion is the combined
\begin{equation*}
    [\sigma_\theta(t, X_t),\, g_\phi(t, H_t) \alpha_\theta \sigma_\theta(t, X_t)].
\end{equation*}
$H_T$ is extracted from the final hidden state, and $m_\phi$ applied, to produce the discriminator's score for that sample.

\paragraph{Dense data regime}
We still need to apply the discriminator to the training data.

First suppose that we observe samples from $Y_{\text{true}}$ as an irregularly sampled time series $\mathbf{z} = ((t_0, z_0), \ldots, (t_n, z_n))$, potentially with missing data, but which is (informally speaking) densely sampled. Without loss of generality let $t_0 = 0$ and $t_n = T$.

Then it is enough to interpolate $\widehat{z} \colon [0, T] \to \reals^y$ such that $\widehat{z}(t_i) = z_i$, and compute
\begin{align}
    H_0 & = \xi_\phi(\widehat{z}(t_0)),\nonumber\\[2pt]
    \dd H_t & = f_\phi(t, H_t) \,\dd t + g_\phi(t, H_t) \circ \dd \widehat{z}_t,\nonumber\\[2pt]
    \qquad D & = m_\phi \cdot H_T,\label{eq:discriminator-data}
\end{align}
where $g_\phi(t, H_t) \circ \dd \widehat{z}_t$ is defined as a Riemann--Stieltjes integral, stochastic integral, or rough integral, depending on the regularity of $\widehat{z}$.

In doing so the interpolation produces a distribution on path space; the one that is desired to be modelled. For example linear interpolation \citep{levin2013}, splines \citep{kidger2020neuralcde}, Gaussian processes \citep{gp-adapter1, gp-adapter2} and so on are all acceptable.

In each case the relatively dense sampling of the data makes the choice of interpolation largely unimportant. We use linear interpolation for three of our four experiments (stocks, air quality, weights) later.

\begin{figure*}[!b]
    \centering
    \includegraphics[width=0.19\textwidth]{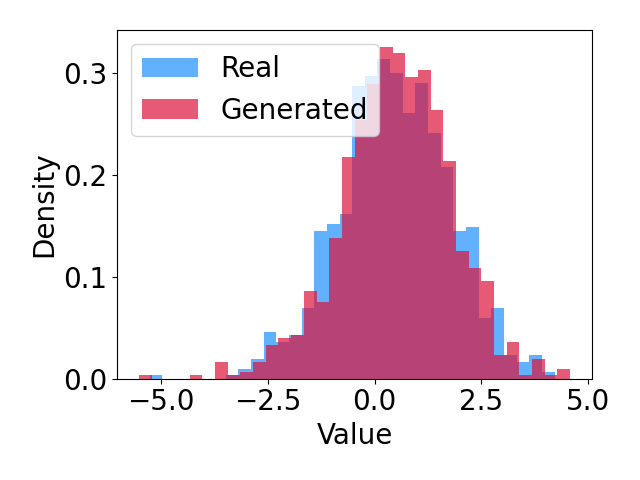}
    \includegraphics[width=0.19\textwidth]{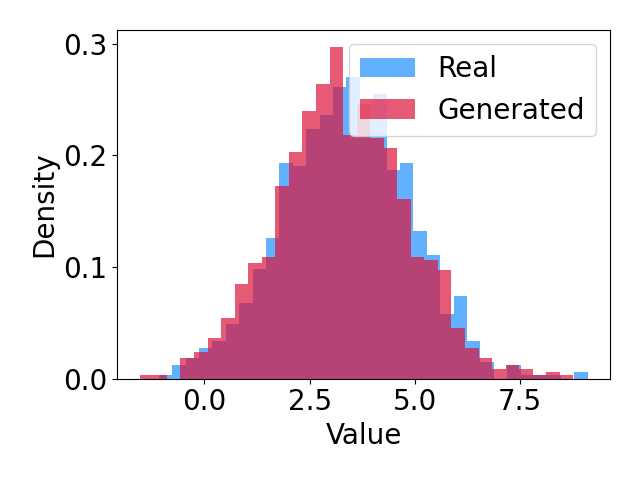}
    \includegraphics[width=0.19\textwidth]{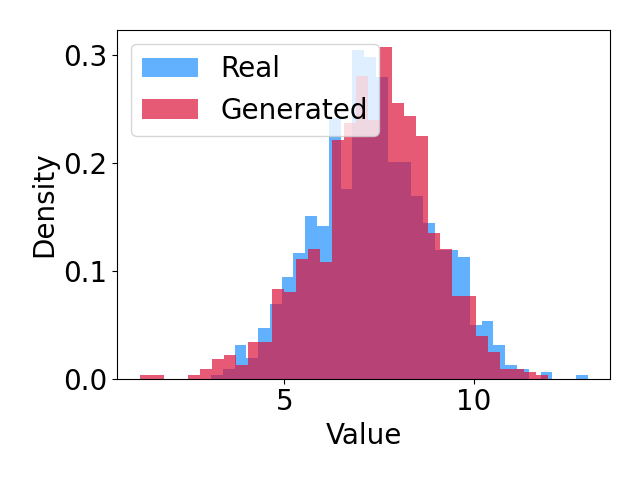}
    \includegraphics[width=0.19\textwidth]{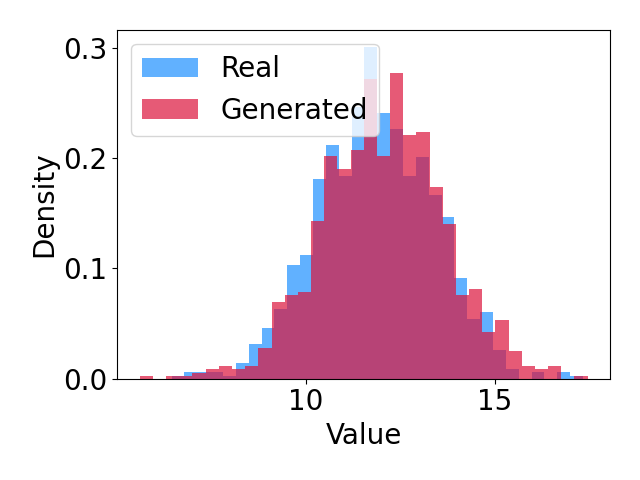}
    \includegraphics[width=0.19\textwidth]{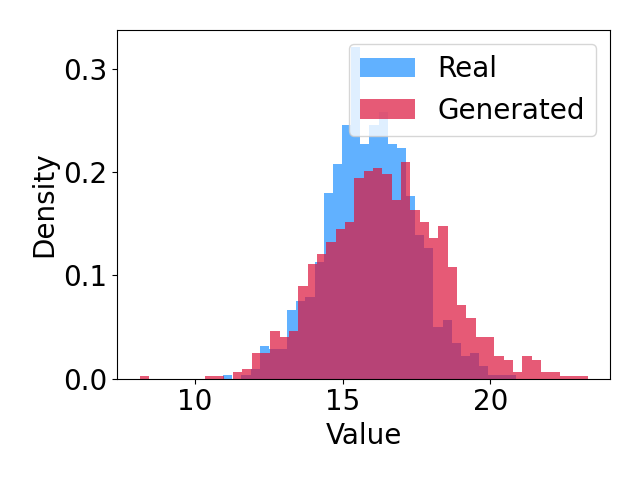}
    \vspace{-1em}
    \caption{Left to right: marginal distributions at $t = 6, 19, 32, 44, 57$.}\label{fig:synthetic-marginal}
\end{figure*}

\paragraph{Sparse data regime}
The previous option becomes a little less convincing when $\mathbf{z}$ is potentially sparsely observed. In this case, we instead first sample the generator at whatever time points are desired, and then interpolate both the training data and the generated data -- solving equation \eqref{eq:discriminator-data} in both cases.

In this case, the choice of interpolation is simply part of the discriminator, and the interpolation is simply a way to embed discrete data into continuous space. We use this approach for the time-dependent Ornstein--Uhlenbeck experiment later.

\paragraph{Training loss}\label{section:loss}
The training losses used are the usual one for Wasserstein GANs \citep{gan, wgan}. Let $Y_\theta \colon (V, W) \mapsto Y$ represent the overall action of the generator, and let $D_\phi \colon Y \mapsto D$ represent the overall action of the discriminator. Then the generator is optimised with respect to
\begin{equation}\label{eq:generator-objective}
    \min_\theta\, \left[ \expect_{V, W} D_\phi(Y_\theta(V, W)) \right],
\end{equation}
and the discriminator is optimised with respect to
\begin{equation}\label{eq:discriminator-objective}
    \max_\phi\, \left[ \expect_{V, W} D_\phi(Y_\theta(V, W)) - \expect_{\mathbf{z}} D_\phi(\widehat{z}) \right].
\end{equation}

Training is performed via stochastic gradient descent techniques as usual.

\paragraph{Lipschitz regularisation}
Wasserstein GANs need a Lipschitz discriminator, for which a variety of methods have been proposed. We use gradient penalty \citep{improved-wgan}, finding that neither weight clipping nor spectral normalisation worked \citep{wgan, miyato2018spectral}.

We attribute this to the observation that neural SDEs (as with RNNs) have a recurrent structure. If a single step has Lipschitz constant $\lambda$, then the Lipschitz constant of the overall neural SDE will be $\bigO(\lambda^T)$ in the time horizon $T$. Even small positive deviations from $\lambda = 1$ may produce large Lipschitz constants. In contrast gradient penalty regularises the Lipschitz constant of the entire discriminator.

Training with gradient penalty implies the need for a double backward. If using the continuous-time adjoint equations of \citep{scalable-sde}, then this implies the need for a double-adjoint. Mathematically this is fine: however for moderate step sizes this produces gradients that are sufficiently inaccurate as to prevent models from training. For this reason we instead backpropagate through the internal operations of the solver.

\paragraph{Learning any SDE} The Wasserstein metric has a unique global minima at $Y=Y_\text{true}$. By universal approximation of Neural CDEs (with respect to either continuous inputs or interpolated sequences, corresponding to dense and sparse data regimes respectively) \citep{kidger2020neuralcde}, the discriminator is sufficiently powerful to approximate the Wasserstein metric over any compact set of inputs.

Meanwhile by the universal approximation theorem for neural networks \citep{pinkus, deepandnarrow} and convergence results for SDEs \citep[Theorem 10.29]{friz-victoir} it is immediate that any (Markov) SDE of the form
\begin{equation*}
    \dd Y_t = \mu(t, Y_t)\, \dd t + \sigma(t, Y_t) \circ \dd W_t
\end{equation*}
may be represented by the generator. Beyond this, the use of hidden state $X$ means that non-Markov dependencies may also be modelled by the generator. (This time without theoretical guarantees, however -- we found that proving a formal statement hit theoretical snags.)







\section{Experiments}
We perform experiments across four datasets; each one is selected to represent a different regime. First is a univariate synthetic example to readily compare model results to the data. Second is a large-scale (14.6 million samples) dataset of Google/Alphabet stocks. Third is a conditional generative problem for air quality data in Beijing. Fourth is a dataset of weight evolution under SGD.

In all cases see Appendix \ref{appendix:experiments} for details of hyperparameters, learning rates, optimisers and so on.

\subsection{Synthetic example: time-dependent Ornstein--Uhlenbeck process}
We begin by considering neural SDEs only (our other experiments feature comparisons to other models), and attempt to mimic a time-dependent one-dimensional Ornstein--Uhlenbeck process. This is an SDE of the form
\begin{equation*}
    \dd z_t = (\mu t - \theta z_t)\, \dd t + \sigma \circ \dd W_t.
\end{equation*}
We let $\mu = 0.02, \theta = 0.1, \sigma = 0.4$, and generate 8192 samples from $t = 0$ to $t = 63$, sampled at every integer.

\paragraph{Marginal distributions}
We plot marginal distributions at $t = 6, 19, 32, 44, 57$. (Corresponding to 10\%, 30\%, 50\%, 70\% and 90\% of the way along.) See Figure \ref{fig:synthetic-marginal}. We can visually confirm that the model has accurately recovered the true marginal distributions. 

\begin{table*}[!b]
\centering
\caption{Results for stocks dataset. Bold indicates best performance; mean $\pm$ standard deviation over three repeats.}\label{table:stocks}
\begin{tabular}{@{}lccc@{}}
\toprule
Metric & Neural SDE & CTFP & Latent ODE \\ \midrule
Classification & \textbf{0.357 $\pm$ 0.045} & 0.165 $\pm$ 0.087 & 0.000239 $\pm$ 0.000086 \\
Prediction & \textbf{0.144 $\pm$ 0.045} & 0.725 $\pm$ 0.233 & 46.2 $\pm$ 12.3 \\
MMD & \textbf{1.92 $\pm$ 0.09} & 2.70 $\pm$ 0.47 & 60.4 $\pm$ 35.8 \\ \bottomrule
\end{tabular}
\end{table*}

\paragraph{Sample paths}
Next we plot 50 samples from the true distribution against 50 samples from the learnt distribution.

\begin{figure}\centering
\includegraphics[width=\linewidth]{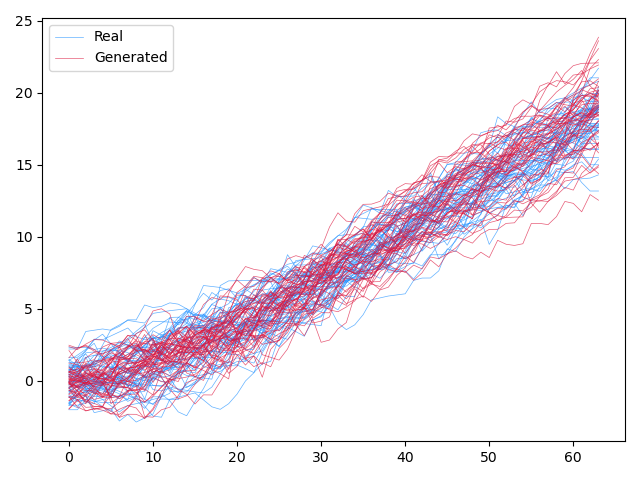}
\vspace{-2em}
\caption{Sample paths from the time-dependent Ornstein--Uhlenbeck SDE, and from the neural SDE trained to match it.}\label{fig:synthetic-sample}
\end{figure}

See Figure \ref{fig:synthetic-sample}. Once again we see excellent agreement between the data and the model.

Overall we see that the neural SDEs are sufficient to recover classical non-neural SDEs: at least on this experiment, nothing has been lost in the generalisation.

\subsection{Google/Alphabet stock prices}
\paragraph{Dataset} Next we consider a dataset consisting of Google/Alphabet stock prices, obtained from LOBSTER \citep{lobster}. The data consists of limit orders, in particular ask and bid prices.

A year of data corresponding to 2018--2019 is used, with an average of 605\,054 observations per day. 
This is then downsampled and sliced into windows of length approximately one minute, for a total of approximately 14.6 million datapoints. We model the two-dimensional path consisting of the midpoint and the log-spread.

\paragraph{Models} Here we compare against two recently-proposed and state-of-the-art competing neural differential equation models; specifically the Latent ODE model of \citet{latent-odes} and the continuous time flow process (CTFP) of \citet{ctfp}. The extended version of CTFPs, including latent variables, is used.

Between them these models cover several training regimes. Latent ODEs are trained as variational autoencoders; CTFPs are trained as normalising flows; neural SDEs are trained as GANs. (To the best of our knowledge neural SDEs as considered here are in fact the first model in their class, namely continuous-time GANs.)

\paragraph{Performance metrics} We study three test metrics: classification, prediction, and MMD.


\emph{Classification} is given by training an auxiliary model to distinguish real data from fake data.  We use a neural CDE \citep{kidger2020neuralcde} for the classifier. Larger losses, meaning inability to classify, indicate better performance of the generative model.

\emph{Prediction} is a train-on-synthetic-test-on-real (TSTR) metric \citep{tstr}. We train a sequence-to-sequence model to predict the latter part of a time series given the first part, using generated data. Testing is performed on real data. We use a neural CDE/ODE as an encoder/decoder pair. Smaller losses, meaning ability to predict, are better.

\emph{Maximum mean discrepancy} is a distance between probability distributions with respect to a kernel or feature map. We use the depth-5 signature transform as the feature map \citep{kiraly2019kernels, toth2019gp}. Smaller values, meaning closer distributions, are better.

\paragraph{Results} The results are shown in Table \ref{table:stocks}. We see that neural SDEs outperform both competitors in all metrics. Notably the Latent ODE fails completely on this dataset. We believe this reflects the fact the stochasticity inherent in the problem; this highlights the inadequacy of neural ODE-based modelling for such tasks, and the need for neural SDE-based modelling instead.


\begin{table*}
\centering
\caption{Results for air quality dataset. Bold indicates best performance; mean $\pm$ standard deviation over three repeats.}\label{table:air-quality}
\begin{tabular}{@{}lccc@{}}
\toprule
Metric & Neural SDE & CTFP & Latent ODE \\ \midrule
Classification & 0.589 $\pm$ 0.051 & \textbf{0.764 $\pm$ 0.064} & 0.392 $\pm$ 0.011 \\
Prediction & \textbf{0.395 $\pm$ 0.056} & 0.810 $\pm$ 0.083 & 0.456 $\pm$ 0.095 \\
MMD & \textbf{0.000160 $\pm$ 0.000029} & 0.00198 $\pm$ 0.00001 & 0.000242 $\pm$ 0.000002 \\ \bottomrule
\end{tabular}
\end{table*}
\begin{table*}
\centering
\caption{Results for weights dataset. Bold indicates best performance; mean $\pm$ standard deviation over three repeats.}\label{table:weights}
\begin{tabular}{@{}lccc@{}}
\toprule
Metric & Neural SDE & CTFP & Latent ODE \\ \midrule
Classification & 0.507 $\pm$ 0.019 & \textbf{0.676 $\pm$ 0.014} & 0.0112 $\pm$ 0.0025 \\
Prediction & \textbf{0.00843 $\pm$ 0.00759} & 0.0808 $\pm$ 0.0514 & 0.127 $\pm$ 0.152 \\
MMD & \textbf{5.28 $\pm$ 1.27} & 12.0 $\pm$ 0.5 & 23.2 $\pm$ 11.8 \\ \bottomrule
\end{tabular}
\end{table*}

\subsection{Air Quality in Beijing}
Next we consider a dataset of the air quality in Beijing, from the UCI repository \citep{beijing, uci}. Each sample is a 6-dimensional time series of the SO$_2$, NO$_2$, CO, O$_3$, PM$_{2.5}$ and PM$_{10}$ 
concentrations, as they change over the course of a day.

We consider the same collection of models and performance statistics as before. We train this as a conditional generative problem, using class labels that correspond to 14 different locations the data was measured at. Class labels are additionally made available to the auxiliary models performing classification and prediction. See Table \ref{table:air-quality}.

On this problem we observe that neural SDEs win on two out of the three metrics (prediction and MMD). CTFPs outperform neural SDEs on classification; however the CTFP severely underperforms on prediction. We believe this reflect the fact that CTFPs are strongly diffusive models; in contrast see how the drift-only Latent ODE performs relatively well on prediction. Once again this highlights the benefits of SDE-based modelling, with its combination of drift and diffusion terms.

\subsection{Weights trained via SGD}
Finally we consider a problem that is classically understood via (stochastic) differential equations: the weight updates when training a neural network via stochastic gradient descent with momentum. We train several small convolutional networks on MNIST \citep{lecun2010mnist} for 100 epochs, and record their weights on every epoch. This produces a dataset of univariate time series; each time series corresponding to a particular scalar weight.

We repeat the comparisons of the previous section. Doing so we obtain the results shown in Table \ref{table:weights}. Neural SDEs once again perform excellently. On this task we observe similar behaviour to the air quality dataset: the CTFP obtains a small edge on the classification metric, but the neural SDE outcompetes it by an order of magnitude on prediction, and by a factor of about two on the MMD. Latent ODEs perform relatively poorly in comparison to both.

\subsection{Successfully training neural SDEs}\label{section:considerations}
As a result of our experiments, we empirically observed that successful training of neural SDEs was predicated on several factors.

\paragraph{Final tanh nonlinearity} Using a final tanh nonlinearity (on both drift and diffusion, for both generator and discriminator) constrains the rate of change of hidden state. This avoids model blow-up as in \citet{kidger2020neuralcde}.

\paragraph{Stochastic weight averaging} Using the Ces{\`a}ro mean of both the generator and discriminator weights, averaged over training, improves performance in the final model \citep{swa}. This averages out the oscillatory training behaviour for the min-max objective used in GAN training.

\paragraph{Adadelta} We experimented with several different standard optimisers, in particular including SGD, Adadelta \cite{zeiler2012adadelta} and Adam \cite{kingma2015}. Amongst all optimisers considered, Adadelta produced substantially better performance. We do not have an explanation for this.

\paragraph{Weight decay} Nonzero weight decay also helped to damp the oscillatory behaviour resulting from the min-max objective used in GAN training.


\section{Conclusion}
By coupling together a neural SDE and a neural CDE as a generator/discriminator pair, we have shown that neural SDEs may be trained as continuous time GANs. Moreover we have shown that this approach extends the existing classical approach to SDE modelling -- using prespecified payoff functions -- so that it may be integrated into existing SDE modelling workflows. Overall, we have demonstrated the capability of neural SDEs as a means of modelling distributions over path space.

\subsubsection*{Acknowledgements}
PK was supported by the EPSRC grant EP/L015811/1. JF was supported by the EPSRC grant EP/N509711/1. PK, JF, HO, TL were supported by the Alan Turing Institute under the EPSRC grant EP/N510129/1. PK thanks Penny Drinkwater for advice on Figure \ref{fig:summary}.

\bibliography{nsde}
\bibliographystyle{icml2021}
\appendix
\clearpage

\section{Experimental details}\label{appendix:experiments}
\subsection{General notes}

\paragraph{Sample code}
A straightforward pedagogical implementation has been made available as an example in the \texttt{torchsde} library \citep{torchsde}.

\paragraph{Software} We used PyTorch \citep{pytorch} as an autodifferentiable framework. We used the \texttt{torchsde} library \citep{torchsde} to solve SDEs. We used the Signatory library \citep{signatory} to calculate the signatures used in the MMD metric. We used the \texttt{torchcde} library \citep{torchcde} for its interpolation schemes, and to solve the neural CDEs used in the classification and prediction metrics. We used the \texttt{torchdiffeq} library \citep{torchdiffeq} to solve the neural ODEs used in the classification and prediction metrics, and for the ODE components of the Latent ODE and CTFP models.

\paragraph{Computing infrastruture} Training was performed on computers using Ubuntu 18.04 LTS, across a mix of five GeForce RTX 2080 Ti and two Quadro GP100; each experiment was performed on a single GPU. Training time varied by problem; none took more than a week per experiment.

\paragraph{SDE solvers} The SDEs used the midpoint method. Recall that the target time series data was regularly sampled and linearly interpolated to make a path. We took the SDE solver to take a single step between each output data point.

\paragraph{ODE solvers} All ODEs were solved using the midpoint method. (For consistency with the solvers used to train the SDE, for example to ensure that the discriminator's action on real data is similar to the discriminator's action on generated data.)

\paragraph{CDE solvers} The CDEs of the classification and prediction models were solved by reducing to ODEs as in \citet{kidger2020neuralcde}.

\paragraph{Normalisation} All data was normalised to have zero mean and unit variance.

\paragraph{Architectures} To recap, the neural SDE has generator initial condition $\zeta_\theta$, generator drift $\mu_\theta$, generator diffusion $\sigma_\theta$, discriminator initial condition $\xi_\phi$, discriminator drift $f_\phi$, and discriminator diffusion $g_\phi$. All of these are parameterised as neural networks.

Meanwhile Latent ODEs have an ODE-RNN encoder (with a neural network vector field) and a neural ODE decoder (with a neural network vector field). The CTFP has an ODE-RNN encoder (with a neural network vector field) and a continuous normalising flow \citep{neural-odes, ffjord} (with a neural network vector field) Additionally \citet{ctfp} condition the normalising flow on the time evolution of a neural ODE of some latent state, which requires another neural network vector field.

Hyperparameters were selected according to informal hyperparameter optimisation across all models.

For the stocks, air quality, and weights datasets (but not the time-dependent Ornstein--Uhlenbeck process, which was done separately and is described below), every neural network was parameterised as a feedforward network with 2 hidden layers, width 64, and softplus activations. The drift, diffusion and vector fields, for every model, all additionally had a tanh nonlinearity as their final operation. As described in the main text we found that this improved the performance of every model.

The neural SDE's generator has hidden state of size $x$ and the discriminator has hidden state is of size $h$. These were both taken as $x = h = 96$. Note that this is larger than the width of each hidden layer within the neural networks, so that the first operation within each neural network is a map from $\reals^{96} \to \reals^{64}$. Somewhat anecdotally, we found that taking the state to be larger than the hidden width was beneficial for model performance.\footnote{This has some loose theoretical justification: a signature is a linear differential equation with very large state, and it is a universal approximator. (See \citet[Appendix B]{kidger2020neuralcde} and references within -- this is a classical fact within rough analysis.) That is to say, it is a simple vector field with a large state, rather than a complicated vector field with a small state.}

The Latent ODE likewise has evolving hidden state, which was also taken to be of size 96.

The Latent ODE samples noise from a normally distributed initial condition, which we took to have 40 dimensions. The CTFP samples noise from a Brownian motion, which as a continuous normalising flow has dimension equal to the number of dimensions of target distribution.

The neural SDE samples noise from both a normally distributed initial condition and a Brownian motion. For the stocks, air quality and weights datasets (but not the time-dependent Ornstein--Uhlenbeck process, which was done separately and is described below), we took the initial condition to have 40 dimensions. The number of dimensions of the Brownian motion was dataset dependent, see below.

The CTFP included a latent context vector as described in \citet{ctfp}. This was taken to have 40 dimensions.

\paragraph{Optimisers} The CTFP and Latent ODE were both trained with Adam \citep{kingma2015} with a learning rate of $4 \times 10^{-5}$. The generator and discriminator of the neural SDE were trained with Adadelta with a learning rate of $4 \times 10^{-5}$. The learning rates were chosen by starting at $4 \times 10^{-4}$ (arbitrarily) and reducing until good performance was achieved. (In particular seeking to avoid oscillatory behaviour in training of the neural SDE.)

\paragraph{Training} For the stocks, air quality and weights datasets (but not the time-dependent Ornstein--Uhlenbeck process, which was done separately and is described below), every model was trained for 100 epochs. The discriminator of the neural SDE received five training steps for every step with which the generator was trained, as is usual; the number of epochs given at 100 is for the generator, for a fair comparison to the other models.

Batch sizes were picked based on what was the largest possible batch size that GPU memory allowed for; these vary by problem and are given below.

\paragraph{Classifier and predictor} The classifier was taken to be a neural CDE with hidden state of size 32, and whose vector field was parameterised as a feedforward neural network with 2 hidden layers of width 32, with softplus activations and final tanh activation.

The predictor was taken to be a neural CDE/neural ODE encoder/decoder pair. Both had a hidden state of size 32, and vector fields parameterised as feedforward neural network with 2 hidden layers of width 32, with softplus activations and final tanh activation. 32 dimensions were used at the encoder/decoder interface.

The learning rate used was $10^{-4}$ for both models, for every dataset and generative model considered, with the one exception of CTFP on Beijing Air Quality, where we observed divergent training of the classifier; the learning rate was reduced to $10^{-5}$ for this case only.

In all cases they were trained for 50 epochs using Adam, with early stopping if the model failed to improve its training loss over 20 epochs.

The classifier took an 80\%/20\% train/test split of the dataset given by combining the underlying dataset and model-generated samples of equal size.

\subsection{Time-dependent Ornstein--Uhlenbeck process}
Each sample is of length 64. The batch size was 1024. The learning rate was $10^{-3}$ The neural SDE was trained for 6000 steps. (Not epochs.) $L^2$ weight decay with scaling 0.01 was applied. Weight averaging (over both generator and discriminator) was performed over the final 5500 steps.

The Brownian motion from which noise was sampled has 3 dimensions. The initial noise was sampled over 5 dimensions. The evolving hidden states were taken to have size 32; the the width of each MLP was taken to be 16; each such MLP had a single hidden layer.

\subsection{Stocks}
Each sample is of length 100.

The batch size was 2048 for every model.

For the neural SDE, the discriminator received 1 epoch of training before the main training (of both generator and discriminator simultaneously) commenced. The weight averaging (over both generator and discriminator) was over every training epoch. The Brownian motion from which noise was sampled had 3 dimensions.

The prediction metric was based on using the first 80\% of the input to predict the last 20\%.

\subsection{Beijing Air Quality}
Each sample is of length 24.

The data was normalised to have zero mean and unit variance.

The batch size was 1024 for every model.

For the neural SDE, the discriminator received 10 epochs of training before the main training (of both generator and discriminator simultaneously) commenced. The weight averaging (over both generator and discriminator) was over the final 40 epochs of training. (We realised that this was an obvious improvement over averaging every epoch, as was done for the previous two experiments.) The Brownian motion from which noise was sampled had 10 dimensions.

The prediction metric was based on using the first 50\% of the input to predict the last 50\%. (An accidental change from the 80\%/20\% split used in the other experiments; this was kept as it is fair, as it is the same for all models on this dataset.)

\subsection{Weights}
Each sample is of length 100. Each sample corresponds to the trajectory of a single scalar weight, epoch-by-epoch, as a small convolutional model is trained on MNIST for 50 epochs. Every weight from the network is used, and treated as a separate sample. This is repeated 10 times. If $P$ is the number of parameters in the convolutional network, then the overall size of the dataset is now $(samples=10P, length=100, channels=1)$.

The batch size was 4096 for the neural SDE and latent ODE. This was reduced to 1024 for the CTFP, which we found to be a very memory intensive model on this problem.

For the neural SDE, the discriminator received 10 epochs of training before the main training (of both generator and discriminator simultaneously) commenced. The weight averaging (over both generator and discriminator) was over every training epoch. The Brownian motion from which noise was sampled had 3 dimensions.

The prediction metric was based on using the first 80\% of the input to predict the last 20\%.
\end{document}